\documentclass{article}
\usepackage{authblk}
\usepackage{listings}
\usepackage{float}
\usepackage{subcaption}
\usepackage{hyperref} 
\usepackage{graphicx} 

\title{Multimodal AI for Gastrointestinal Diagnostics: Tackling VQA in MEDVQA-GI 2025}

\author[1]{Sujata Gaihre}
\author[1]{Amir Thapa Magar}
\author[2]{Prasuna Pokharel}
\author[3]{Laxmi Tiwari}

\affil[1]{NCIT, Nepal}
\affil[2]{Fusemachine, Nepal}
\affil[3]{Logictronix Technologies, Nepal}

\date{}

\begin{document}

\maketitle

\begin{abstract}
This paper describes our approach to Subtask~1 of the ImageCLEFmed MEDVQA~2025 Challenge, which targets visual question answering (VQA) for gastrointestinal endoscopy. We adopt the Florence model---a large-scale multimodal foundation model---as the backbone of our VQA pipeline, pairing a powerful vision encoder with a text encoder to interpret endoscopic images and produce clinically relevant answers. To improve generalization, we apply domain-specific augmentations that preserve medical features while increasing training diversity. Experiments on the KASVIR dataset show that fine-tuning Florence yields accurate responses on the official challenge metrics. Our results highlight the potential of large multimodal models in medical VQA and provide a strong baseline for future work on explainability, robustness, and clinical integration. The code is publicly available at: \href{https://github.com/TiwariLaxuu/VQA-Florence.git}{github.com/TiwariLaxuu/VQA-Florence.git}.

\end{abstract}
\textbf{Keywords:} Medical VQA, ImageCLEFmed 2025, Multimodal AI, Clinical Question Answering
\section{Introduction}

Early detection and treatment of gastrointestinal (GI) diseases depend on the accurate interpretation of endoscopic images. Recent advances in deep learning offer promising solutions to automate this analysis, enabling a timely and reliable diagnosis. Visual Question Answering (VQA) further enhances these systems by linking image understanding with natural language queries to provide actionable clinical insights~\cite{Gautam2023Oct, Gautam2025Jun}. Synthetic image generation also plays a key role by expanding training data without the need for extensive manual annotation.

The ImageCLEFmed MEDVQA 2025 challenge encourages progress in this domain, with Subtask~1 dedicated to VQA for GI endoscopy. We present a reproducible pipeline that leverages a multimodal foundation model to answer clinically relevant questions. Our approach demonstrates that careful data augmentation and streamlined fine-tuning can yield accurate results while remaining accessible to the research community.

\section{Related Work}
VQA has evolved significantly over the past decade, with growing emphasis on reducing dataset biases and ensuring visual grounding in answers~\cite{zakari2022vqa}. Early VQA models demonstrated high performance by exploiting statistical patterns in the questions, rather than truly interpreting the content of the image - a critical flaw when applying such models to sensitive domains such as medical diagnostics.

Medical Visual Question Answering (Med-VQA) has rapidly evolved with advances in both computer vision and natural language processing. Traditional Med-VQA approaches include Modality-Ensemble Visual Features (MEVF)—which integrate visual cues across modalities—combined with Bilinear Attention Networks (BAN) a technique for modeling image-question interactions using low-rank bilinear pooling. Conditional Reasoning (CR)~\cite{zhan2020medical}, as well as Contrastive Pretraining and Representation Distillation (CPRD) with BAN~\cite{10.1007/978-3-030-87196-3_20}, treated the problem as a classification task, relying heavily on predefined answer sets. These methods struggled with open-ended questions due to limited integration of external medical knowledge and semantic reasoning.

More recent work has explored visual language pretraining. PubMedCLIP ~\cite{eslami2023pubmedclip} leveraged contrastive learning in medical text image pairs, while Masked Multimodal Masked Autoencoder (M3AE) ~\cite{CHEN2024103018} used masked modeling for joint vision-language alignment. The state-of-the-art multimodal concept alignment pre-training (MMCAP) ~\cite{yan-etal-2024-multi} proposed a novel Multi-modal Concept Alignment Pre-training approach using a knowledge graph derived from the Unified Medical Language System (UMLS)  and image-caption datasets. By aligning visual and textual data through a transformer-based encoder-decoder framework with external medical knowledge, MMCAP achieved top performance on both Semantically-Labeled Knowledge-Enhanced Dataset(SLAKE)~\cite{liu2021slake}and VQA on radiology image datasets~\cite{lau2018dataset}.

A pivotal work addressing this issue was proposed by Goyal et al.~\cite{goyal2017making} in "Making the V in VQA Matter". The authors identified that models trained on the existing VQA dataset~\cite{antol2015vqa} often relied heavily on language priors. For example, questions like "Is there a clock?" could be correctly answered without analyzing the image due to dataset bias. To address this, they introduced VQA v2.0~\cite{goyal2017making}, a balanced dataset that pairs each question with two visually similar images requiring different answers. This structure significantly reduced reliance on question-only cues, forcing models to ground their answers in visual content. Their findings showed a noticeable performance drop in models previously successful on original VQA, confirming the overreliance on language. They also proposed a counter-example retrieval method as a basic form of model interpretability. These design principles—bias mitigation, dataset balancing, and explainability—are highly relevant to medical VQA, where clinical safety depends on faithful visual reasoning.

Building on these foundational ideas, Gautam et al.~\cite{KvasirVQA} introduced Kvasir-VQA, a domain-specific VQA dataset for GI endoscopy. Derived from the HyperKvasir dataset, Kvasir-VQA consists of over 6,500 annotated image-question-answer (IQA) triplets, with questions closely aligned to real clinical scenarios. Unlike generic VQA datasets, Kvasir-VQA emphasizes clinically significant questions across diverse GI conditions, procedures, and anatomical regions. This design enables models to learn nuanced multimodal patterns critical for accurate diagnostic reasoning. To address the scarcity of annotated medical images, the dataset integrates expert-verified questions spanning identification, localization, and reasoning tasks—thereby advancing medical AI benchmarks in the GI domain.

While earlier approaches on Kvasir-VQA have utilized baseline multimodal models with standard data augmentation, our approach for MEDVQA 2025 Task 1 builds upon and extends this line of work. We adopt Florence, a large-scale multimodal transformer known for robust visual-language alignment, and introduce domain-specific image augmentations tailored for endoscopic imagery. These augmentations preserve critical visual features (e.g., mucosal texture, bleeding points) while simulating real-world variability. Additionally, we employ a generative decoding strategy that enables our model to produce clinically precise, open-ended answers, in contrast to classification-based systems that limit expressiveness. These innovations lead to state-of-the-art performance on the Kvasir-VQA benchmark and represent a promising step toward trustworthy medical VQA systems.

Guo et al.~\cite{guo2024unk}  tackled the often-overlooked problem of unanswerable questions in VQA. In real-world applications, including clinical and scientific settings, VQA systems frequently encounter questions that cannot be answered given the provided visual input. However, most existing VQA benchmarks fail to account for these scenarios, leading models to produce confident yet incorrect answers that can be misleading or even harmful.

To address this, the authors introduced unknown visual question answering (UNK-VQA), a dataset specifically designed to evaluate a model’s ability to recognize when a question is unanswerable. They constructed this dataset by systematically modifying answerable questions from standard VQA datasets using perturbation techniques such as word replacement, semantic negation, and object substitution. These modifications generated challenging questions that remained linguistically coherent but lacked sufficient visual evidence to answer correctly. By combining both answerable and unanswerable questions, UNK-VQA provides a rigorous test of a model’s abstention capabilities.

Guo et al. conducted extensive evaluations using several state-of-the-art vision-language models, including BLIP ~\cite{li2022blip}, LLaVA~\cite{liu2024improved}, and GPT-4V~\cite{guo2024unk}. Despite performing well on traditional VQA benchmarks, these models often failed on UNK-VQA, frequently producing overconfident yet incorrect answers. This finding highlights a significant limitation in current VQA architectures: the inability to reliably abstain from answering when faced with insufficient visual information.

Overall, UNK-VQA offers a valuable resource for the community to evaluate and improve VQA models’ ability to handle uncertainty, a critical requirement for deploying such systems in sensitive domains where incorrect answers can have serious consequences.

In summary, by combining principles from general VQA (e.g., dataset balancing and grounding from Goyal et al.) with domain-specific insights from Kvasir-VQA, our method addresses the unique challenges of medical VQA—ensuring accuracy, interpretability, and clinical relevance in gastrointestinal diagnostics.

\section{Task Description}

In this study, we participated in Subtask 1: Visual Question Answering (VQA) of the ImageCLEFmed-MEDVQA-GI 2025 challenge~\cite{OverviewImageCLEF2025}. The objective of this subtask is to develop intelligent systems capable of automatically answering clinically relevant questions based on GI images. This task is especially important in the medical field, where accurate image-based question answering can support clinical diagnosis, documentation, and education.

\subsection{Dataset Description}

We used the Kvasir-VQA dataset~\cite{KvasirVQA} for developing and evaluating our visual question answering (VQA) models. This multimodal dataset is derived from the extended HyperKvasir image repository and comprises approximately 58,849 image–question–answer (IQA) triplets associated with 6,500 high-resolution gastrointestinal (GI) endoscopy images.

Each image in the dataset is linked to multiple QA pairs and is annotated with detail clinical context. Specifically, each IQA sample includes:

\begin{itemize}
    \item \textbf{Image:} A gastrointestinal (GI) endoscopic image.
    \item \textbf{Source:} The clinical label associated with the image, selected from six predefined categories.
    \item \textbf{Question:} A natural language query pertaining to the image, focusing on diagnostic, anatomical, or procedural aspects.
    \item \textbf{Answer:} A concise response that directly addresses the question.
\end{itemize}

\begin{figure}[htbp]
\centering
\begin{tabular}{cc}
\includegraphics[width=0.4\textwidth]{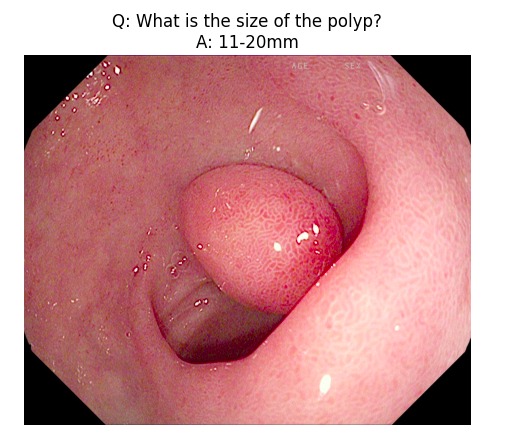} &
\includegraphics[width=0.4\textwidth]{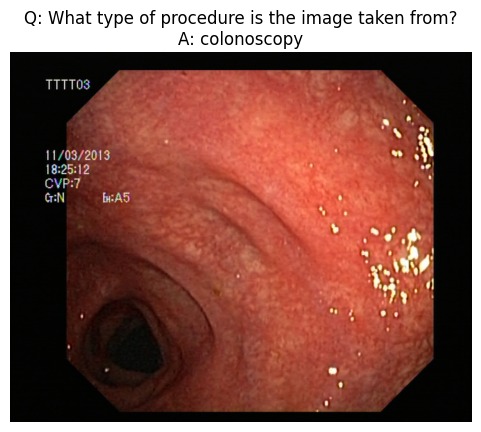} \\
\small (a) Image–question–answer triplet 1 &
\small (b) Image–question–answer triplet 2 \\
\end{tabular}
\caption{Examples of image–question–answer triplets from the Kvasir-VQA dataset.}
\label{fig:sample_triplets}
\end{figure}

For computational efficiency, we utilized a 1\% stratified subset of the full dataset. This subset was initially divided into 90\% training and 10\% testing. The testing portion was further split into 90\% training and 10\% validation, resulting in a final train/validation/test split. This ensured a balanced and representative subset while allowing efficient fine-tuning and evaluation of our models.

\subsection{Exploratory Data Analysis (EDA)}

To better understand the dataset, we conducted a preliminary exploratory data analysis~\cite{chatfield1986exploratory} focusing on the structure and distribution of samples.

\begin{figure}[htbp]
    \centering
    \begin{subfigure}[b]{0.48\textwidth}
        \centering
        \includegraphics[width=\textwidth]{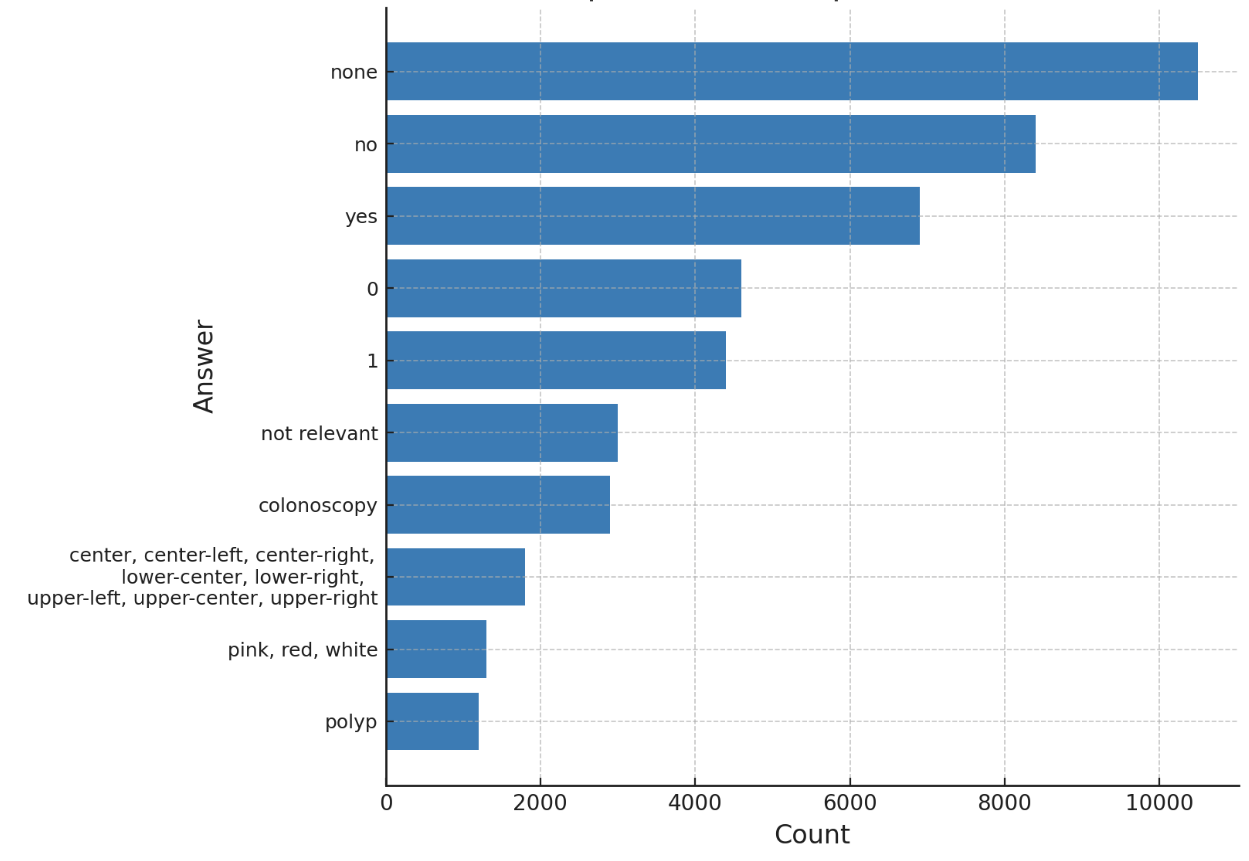}
        \caption{Top 10 most frequent answers}
        \label{fig:top_answers}
    \end{subfigure}
    \hfill
    \begin{subfigure}[b]{0.48\textwidth}
        \centering
        \includegraphics[width=\textwidth]{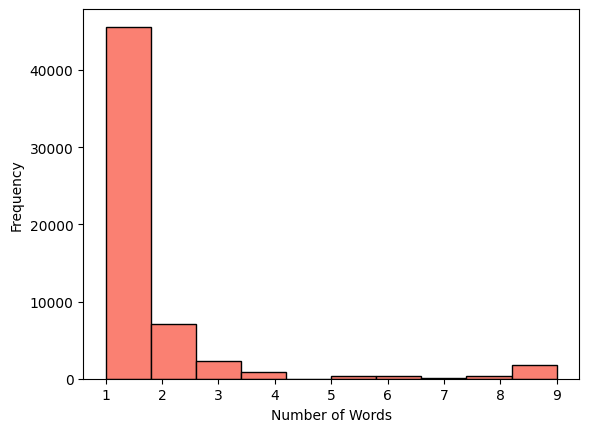}
        \caption{Answer length distribution}
        \label{fig:answer_length}
    \end{subfigure}
    
    \caption{Exploratory data visualizations from the Kvasir-VQA dataset}
    \label{fig:eda_visuals}
\end{figure}

We  visualized  sample data to inspect the quality and diversity of the visual-question-answer triplets. Figure~\ref{fig:sample_triplets} shows examples of endoscopic images with their corresponding clinical questions and answers. These samples reflect a broad spectrum of question types, including disease identification, anatomical assessment, and procedural inquiry.

In addition to qualitative inspection, we quantitatively analyzed the distribution of answers across the dataset. We found that the dataset contains a mix of common and rare answers. Figure~\ref{fig:top_answers} illustrates the top 10 most frequent answers. Short and generic responses such as \textit{none}, \textit{no}, \textit{yes}, and \textit{0} dominate the distribution. This highlights a significant class imbalance, where frequently occurring answers may bias the model if not handled properly during training. Some clinically specific answers like \textit{colonoscopy} and \textit{polyp} are also present, but less frequent.

Figure~\ref{fig:answer_length} shows the distribution of answer lengths measured by the number of words. The majority of answers consist of a single word, with the frequency dropping sharply for longer responses. This indicates that most answers in the dataset are concise and classification-like, rather than descriptive. However, the presence of multi-word answers suggests the need for the model to also handle more complex, free-form responses.

In total, the dataset comprises 58,849 image-question-answer (IQA) samples, based on 20 unique question templates and 502 unique answers. This demonstrates the diversity and complexity of the task, requiring models capable of handling high class imbalance, short-form predictions, and clinically rich semantics across a broad answer space.

\section{Methodology}

To address the challenge of answering clinically relevant questions from gastrointestinal images, we adopt \textbf{Florence-2}—a unified vision foundation model~\cite{xiao2024florence}—as the backbone of our VQA pipeline.

\subsection{Base Model Overview}

Florence-2 supports a wide range of computer vision and vision-language tasks, including image captioning, object detection, referring segmentation~\cite{ding2021vision}, and Visual Question Answering (VQA), using a unified architecture and shared weights. It formulates all tasks within a \textit{sequence-to-sequence} framework: for VQA, the model takes an image and a task-specific prompt (the question) and generates a free-text answer. This prompt-based approach enables consistent inference across modalities and tasks~\cite{jin2021good}.

Florence-2 captures both semantic and spatial detail, which is important in the medical VQA where global (e.g., anatomical site) and local features (e.g., mucosal patterns) inform clinical reasoning. While the model supports spatial grounding through \textit{location tokens}, these were not utilized during fine-tuning due to the lack of region annotations in our dataset.

The unified, prompt-driven design of Florence-2 offers:

\begin{itemize}
    \item \textbf{Flexible answer generation:} Enables detailed, free-form clinical responses beyond classification.
    \item \textbf{Interpretability:} Spatial grounding via location tokens enhances transparency.
    \item \textbf{Scalability:} Pretraining on FLD-5B (5.4B annotations on 126M images) supports generalization to data-scarce domains.
\end{itemize}

These capabilities make Florence-2 particularly suitable for medical VQA tasks such as those in the MEDVQA-GI 2025 challenge.

\subsection{Model Architecture}

\begin{figure}[h]
    \centering
    \includegraphics[width=0.7\textwidth]{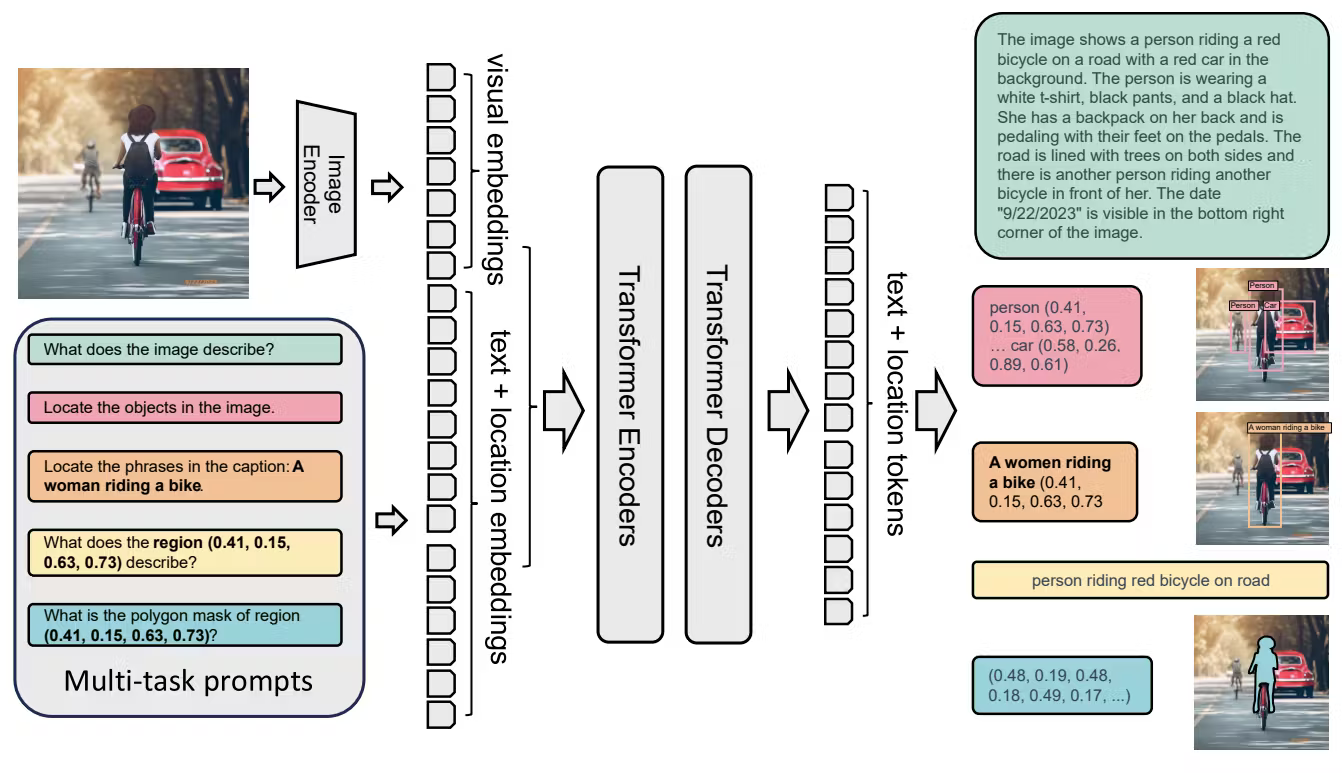}
    \caption{Architecture of Florence-2~\cite{xiao2024florence}}
    \label{fig:samples}
\end{figure}

Florence-2 adopts a modular architecture that integrates a vision encoder and a multi-modal encoder-decoder to align image and text representations. The vision encoder is based on DaViT (Dual Attention Vision Transformer), utilizing a frozen ViT-L/14 backbone pretrained at 896$\times$896 resolution with a patch size of 16$\times$16, resulting in 196 visual tokens per image. These are mapped into a 196$\times$1024 feature map, maintaining spatial information via learned 2D positional embeddings. All parameters of the vision encoder are frozen during fine-tuning to preserve the general-purpose representations from pretraining.

The multi-modal encoder-decoder module fuses visual and textual inputs. Textual prompts are tokenized and embedded, while visual tokens are projected and normalized to match the text embedding space. The concatenated sequence is then processed through a multi-modal encoder to learn joint representations.

For decoding, Florence-2 employs a 2.7B parameter causal language model with 32 transformer layers and 32 attention heads, each with a hidden size of 2048. Cross-attention layers are added every fourth layer to incorporate visual context into the text generation process. During inference, the decoder generates answers step-by-step using temperature sampling ($T = 0.7$), with its attention mechanism using hidden states as queries and the projected 196$\times$256 image features as key-value pairs.

The model is trained using a standard cross-entropy loss objective:
\begin{equation}
\mathcal{L} = - \sum_{i=1}^{|y|} \log P_\theta (y_i \mid y_{<i}, x)
\end{equation}
where $\theta$ represents model parameters, $x$ is the combined input (image and question), and $y$ is the target answer sequence.

\subsection{VQA Adaptability and Fine-Tuning}

Florence-2 performs well in both zero-shot and fine-tuned scenarios:

\begin{itemize}
    \item \textbf{Zero-shot generalization:} Exhibits robust performance without VQA-specific training~\cite{xing2024survey}.
    \item \textbf{Fine-tuning transferability:} Performance improves significantly when adapted to VQA datasets like DocVQA.
    \item \textbf{Unified modeling:} Its prompt-based, task-agnostic formulation eliminates the need for task-specific heads, improving generalizability.
    \item \textbf{Parameter efficiency:} With 0.23B (base) and 0.77B (large) parameters, Florence-2 achieves competitive VQA performance after fine-tuning.
\end{itemize}

\subsection{Fine-Tuning Setup}

We fine-tuned the \texttt{microsoft/Florence-2-base-ft} checkpoint, keeping the ViT-L/14 vision tower frozen. Inputs (images and questions) were processed using the model’s \texttt{AutoProcessor}. Answers were tokenized with padding replaced by \texttt{-100} for causal loss computation. No adapter-based methods (e.g., LoRA) were used. Evaluation employed BLEU, METEOR, and ROUGE-L using the HuggingFace \texttt{evaluate} library.

\subsection{Training Protocol}

Training was conducted using AdamW ($\beta_1=0.9$, $\beta_2=0.999$), a learning rate of $7.8 \times 10^{-6}$ (cosine decay over 20 epochs), and weight decay of 0.1. The effective batch size was 20 (5 per GPU with 4 gradient accumulation steps). Regularization included gradient clipping (max norm = 1.0) and dropout ($p = 0.1$) on attention weights. Evaluation was done after each epoch, with early stopping after 3 epochs without improvement. Baselines were compared using paired t-tests ($p < 0.05$).

\subsection{Implementation Details}

Experiments were implemented in PyTorch with CUDA and HuggingFace Transformers. Training used NVIDIA T4 GPUs (16GB), mixed precision (FP16 for matrix ops, FP32 for loss), and a 72-hour time budget over 10 epochs. Experiment tracking was done using Weights \& Biases, with data/versioning managed by DVC. Reproducibility was ensured through fixed random seeds (42 for data, 3407 for model) and deterministic algorithms.

\section{Results and Evaluation}

In this section, we present the results of our experiments, the evaluation methodology employed, and key insights derived from the analysis. We evaluated the VQA performance using standard NLP metrics including BLEU, ROUGE-1, ROUGE-2, ROUGE-L, and METEOR.

\vspace{0.2em}
\noindent
\textbf{Clarification:} The results reported in Table~\ref{tab:ablation} were obtained after fine-tuning the Florence V2 model on a 1\% stratified subset of the Kvasir-VQA dataset. The model was fine-tuned using domain-specific augmentations and a causal language modeling loss. The vision backbone (ViT-L/14) was frozen during training to preserve pretrained visual features, while the language decoder was trained with the AdamW optimizer (learning rate: $7.8 \times 10^{-6}$, weight decay: 0.1). The training process used a batch size of 5 with gradient checkpointing enabled, and evaluations were performed at each epoch using BLEU, METEOR, and ROUGE-L as primary metrics.

Table~\ref{tab:ablation} summarizes the VQA performance of our fine-tuned model across different evaluation stages. On the validation set, the model achieved a BLEU score of 0.12, ROUGE-1 of 0.78, ROUGE-2 of 0.09, ROUGE-L of 0.77, and a METEOR score of 0.42. Public test results showed improved performance with a BLEU score of 0.150 and a METEOR score of 0.440. The model performed best on the private test set with BLEU 0.160, ROUGE-L 0.880, and METEOR 0.490. 
\begin{table}[ht]
\centering
\caption{Ablation results on Task 1 (VQA) using different augmentations and finetuned Florence V2.}
\begin{tabular}{lccccc}
\hline
\textbf{Set} & \textbf{BLEU} & \textbf{ROUGE-1} & \textbf{ROUGE-2} & \textbf{ROUGE-L} & \textbf{METEOR} \\
\hline
Validation & 0.12 & 0.78 & 0.09 & 0.77 & 0.42 \\
Public     & 0.150 & 0.810 & 0.100 & 0.800 & 0.440 \\
Private    & 0.160 & 0.880 & 0.100 & 0.880 & 0.490 \\
\hline
\end{tabular}
\label{tab:ablation}
\end{table}
\section{Ablation Studies}
In our ablation study, we evaluated four augmentation strategies—no augmentation, heavy, standard, and fine-tuned—using the Florence V2 model for medical VQA. Table 2: Comparison of augmentation strategies and their impact on VQA performance using Florence V2.The baseline (no augmentation) produced low scores (BLEU: 0.00, ROUGE-L: 0.63, METEOR: 0.31), while heavy augmentation further degraded performance due to unrealistic distortions (e.g., vertical flip), with ROUGE-L and METEOR dropping to 0.48 and 0.25, respectively. In contrast, standard augmentation (random crop, flip, color jitter) improved scores (BLEU: 0.12, ROUGE-L: 0.77, METEOR: 0.42). The best results were achieved using fine-tuned augmentation (BLEU: 0.15, ROUGE-L: 0.80, METEOR: 0.44), confirming that carefully tuned, domain-aware transformations enhance model generalization and robustness.

\begin{table}[ht]
\centering
\caption{Impact of augmentation strategies on VQA performance using Florence V2. ROUGE metrics are abbreviated as R-1 (ROUGE-1), R-2 (ROUGE-2), and R-L (ROUGE-L).}
\begin{tabular}{p{4.5cm}ccccc}
\hline
\textbf{Augmentation Strategy} & \textbf{BLEU} & \textbf{R-1} & \textbf{R-2} & \textbf{R-L} & \textbf{METEOR} \\
\hline
No Augmentation              & 0.00 & 0.63 & 0.02 & 0.63 & 0.31 \\
\hline
Heavy\\Augmentation$^{\dagger\dagger}$   & 0.00 & 0.48 & 0.05 & 0.48 & 0.25 \\
\hline
Standard\\Augmentation$^{\dagger}$      & 0.12 & 0.78 & 0.09 & 0.77 & 0.42 \\
\hline
Fine-Tuned\\Augmentation               & 0.15 & 0.81 & 0.10 & 0.80 & 0.44 \\
\hline
\end{tabular}
\label{tab:augmentation}
\end{table}

\vspace{0.2cm}
\noindent
\textbf{*} Standard Augmentation includes random crop, flip, and color jitter.\\
\textbf{**} Heavy Augmentation includes strong distortions like vertical flip and random rotation.

\subsection{Performance by Question Type}

To better understand model behavior~\cite{PromptToPolyp}, we conducted a fine-grained evaluation based on the type of question (e.g., \textit{what}, \textit{where}, \textit{how}). Table~\ref{tab:question_type} reports BLEU, ROUGE-L, and METEOR scores computed separately for each first-word question type in the validation set. This analysis highlights where the model performs well (e.g., ``where'' and ``have'' questions) and where it struggles (e.g., ``how'' and ``is''), which helps us understand where improvements are needed.
\begin{table}[ht]
\centering
\caption{Evaluation metrics by question type on the validation set.}
\begin{tabular}{lccc}
\hline
\textbf{Question Type} & \textbf{BLEU} & \textbf{ROUGE-L} & \textbf{METEOR} \\
\hline
what    & 0.20 & 0.88 & 0.47 \\
is      & 0.00 & 0.87 & 0.44 \\
where   & 0.73 & 0.85 & 0.58 \\
how     & 0.00 & 0.72 & 0.37 \\
have    & 0.00 & 0.91 & 0.77 \\
are     & 0.00 & 0.90 & 0.44 \\
does    & 0.00 & 1.00 & 0.50 \\
\hline
\end{tabular}
\label{tab:question_type}
\end{table}

\section{Discussion}

Our results show that Florence-2, when fine-tuned with clinically informed augmentations, offers a promising baseline for medical VQA in gastrointestinal endoscopy. By freezing the ViT-L/14 vision backbone, we retained robust pretrained features while adapting the decoder to align with domain-specific linguistic patterns.

Among the augmentation strategies evaluated, fine-tuned augmentations provided consistent improvements over both the baseline (BLEU: 0.00, METEOR: 0.31) and heavy augmentations (METEOR: 0.25), achieving the best scores (BLEU: 0.15, METEOR: 0.44). This confirms that medically plausible transformations help preserve critical visual cues necessary for accurate predictions.

Analysis by question type revealed stronger performance on spatial and binary questions—such as those starting with ``where'' (METEOR: 0.58) and ``have'' (0.77)—while procedural and abstract questions like ``how'' proved more difficult (METEOR: 0.37). Additionally, low BLEU scores across categories suggest the model often captures the semantic intent but deviates from exact phrasing—a known limitation of generative models evaluated with n-gram-based metrics.

Finally, performance gains from validation to private test splits (e.g., METEOR: 0.42 $\rightarrow$ 0.49) suggest some generalization, although results remain moderate overall. These findings highlight both the potential and limitations of large multimodal models in specialized clinical VQA tasks, particularly under constrained data conditions.

\section{Conclusion and Future Work}

This study explored the use of Florence-2, a large-scale multimodal foundation model, for visual question answering in gastrointestinal endoscopy. The model was adapted using a frozen ViT-L/14 vision encoder and fine-tuned multimodal layers, showing that even with limited data, clinically meaningful responses can be generated when supported by realistic, domain-specific augmentations.

Fine-tuned augmentation strategies led to notable improvements in performance, with METEOR scores increasing from 0.31 (no augmentation) to 0.44, and up to 0.49 on the private test set. Performance varied by question type—stronger on spatial and binary queries (e.g., ``where,'' ``have'') and weaker on procedural or abstract ones (e.g., ``how'')—indicating strengths in visual pattern recognition but it still struggles with complex clinical reasoning. 

Several directions appear promising for improving clinical applicability: enhancing model interpretability through visual grounding~\cite{hasanpour2022unboxing}, incorporating uncertainty handling for unanswerable cases, enriching semantic reasoning via medical knowledge integration, and extending to multi-turn, conversational scenarios. These steps can improve the system's reliability, transparency, and alignment with real-world clinical workflows.

\section{Acknowledgments}
We sincerely thank the KASVIR dataset providers for their essential contribution to GI diagnostics research. We appreciate the ImageCLEFmed MEDVQA 2025 Challenge organizers for fostering innovation and providing valuable resources. Our gratitude extends to Simula Research Laboratory and contributors to the ImageCLEFmed GitHub repository for their open-source tools that aided our implementation. We also thank Logictronix Technologies for providing computing resources crucial for model training. 
Due to limited access to large-scale infrastructure and datasets, as is often the case in low- and middle-income countries (LMIC) settings, our experiments were conducted on a small subset of the full dataset. We recognize this limitation and hope our work contributes toward greater inclusivity and resource-aware research in the medical AI community.
Lastly, we acknowledge the broader open-source community for their indispensable support in advancing medical AI research.

\bibliographystyle{plainurl}
\bibliography{references}


\end{document}